\documentclass[sigconf]{acmart}

\usepackage{booktabs} 
\usepackage{libertine} 

\usepackage{balance}

\setcopyright{rightsretained}

\acmDOI{}

\acmISBN{}

\acmArticle{4}
\acmPrice{15.00}

\begin{document}
\title{DIF : Dataset of Perceived Intoxicated Faces for Drunk Person Identification}

\author{Vineet Mehta*}
\affiliation{
  \institution{Indian Institute of Technology Ropar}
\authornote{Equal contribution}
}
\email{2016csb1063@iitrpr.ac.in}

\author{Devendra Pratap Yadav*}
\affiliation{
  \institution{Indian Institute of Technology Ropar}
}
\email{2014csb1010@iitrpr.ac.in}

\author{Sai Srinadhu Katta*}
\affiliation{
  \institution{Indian Institute of Technology Ropar}
}
\email{2015csb1015@iitrpr.ac.in}

\author{Abhinav Dhall}
\affiliation{
  \institution{Monash University,\\Indian Institute of Technology Ropar}
}
\email{abhinav.dhall@monash.edu}

\renewcommand{\shortauthors}{}
\begin{abstract}
Traffic accidents cause over a million deaths every year, of which a large fraction is attributed to drunk driving. An automated intoxicated driver detection system in vehicles will be useful in reducing accidents and related financial costs. Existing solutions require special equipment such as electrocardiogram, infrared cameras or breathalyzers. In this work, we propose a new dataset called DIF (Dataset of perceived Intoxicated Faces) which contains audio-visual data of intoxicated and sober people obtained from online sources. To the best of our knowledge, this is the first work for automatic bimodal non-invasive intoxication detection. Convolutional Neural Networks (CNN) and Deep Neural Networks (DNN) are trained for computing the video and audio baselines, respectively. 3D CNN is used to exploit the Spatio-temporal changes in the video. A simple variation of the traditional 3D convolution block is proposed based on inducing non-linearity between the spatial and temporal channels. Extensive experiments are performed to validate the approach and baselines. 
\end{abstract}


\maketitle

\section{Introduction}
Alcohol-impaired driving poses a serious threat to drivers as well as pedestrians. Over 1.25 million road traffic deaths occur every year. Traffic accidents are the leading cause of death among those aged 15-29 years ~\cite{c32}. Drunk driving is responsible for around 40\% of all traffic crashes ~\cite{c33}. Strictly enforcing drunk driving laws can reduce the number of road deaths by 20\% ~\cite{c32}. Modern vehicles are being developed with a focus on smart and automated features. Driver monitoring systems such as drowsiness detectors ~\cite{c34} and driver attention monitors ~\cite{c35} have been developed to increase the safety. In these systems, incorporating an automated drunk detection system is necessary to reduce traffic accidents and related financial costs.

Intoxication detection systems can be divided into three categories :\\ 
1. \textbf{Direct detection} - Measuring Blood Alcohol Content (BAC) directly through breath analysis.\\
2. \textbf{Behavior-based detection} - Detecting characteristic changes in behavior due to alcohol consumption. This may include changes in speech, gait, or facial expressions.\\
3. \textbf{Biosignal-based detection} - Using Electrocardiogram signals ~\cite{c1} or face thermal images ~\cite{c2} to detect intoxication.

\begin{figure}[t]
\centering
\vspace{6mm}
\includegraphics[width=\linewidth]{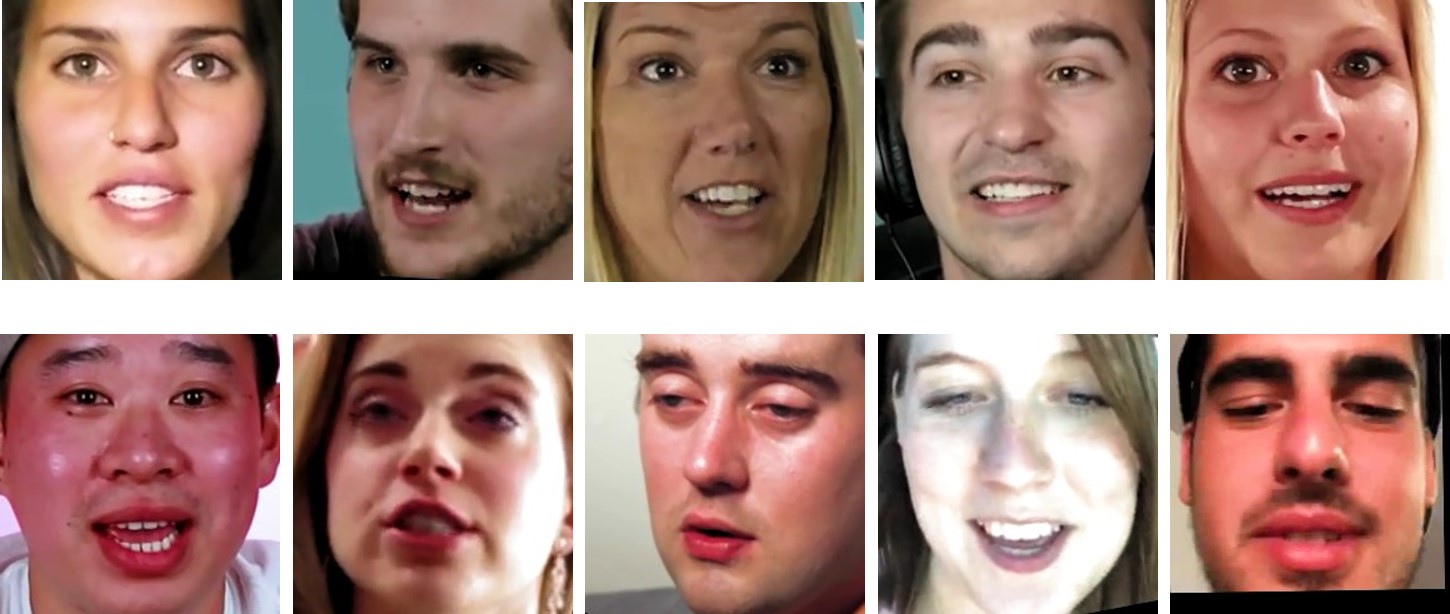}
\caption{Frames from the proposed DIF dataset. The top row contains sober subjects and the bottom row contains intoxicated subjects, respectively.}

\label{DIFFrames}
\end{figure}
Direct detection is often done manually by law enforcement officers using breathalyzers. Biosignal based detection also requires specialized equipment to measure signals. In contrast, behavior-based detection can be performed passively by recording speech or video of the subject and analyzing it to detect intoxication. We focus on behavior-based detection, specifically using face videos of a person. While speech is a widely studied behavioral change in drunk people \cite{maity2018understanding}, we can also detect changes in facial movement. The key-contributions of the paper are as follows:

\begin{itemize}
    \item To the best of our knowledge,  DIF (Dataset of perceived Intoxicated Faces) is the first audio-visual database for learning systems, which can predict if a person is intoxicated or not.
    \item The size of the database is large enough to train deep learning algorithms.
    \item We propose simple changes in the 3D convolutional block's architecture, which improves the classification performance and decreases the number of parameters.
\end{itemize} 

Studies have shown that alcohol consumption leads to changes in facial expressions ~\cite{c36} and significant differences in eye movement patterns ~\cite{c37}. Drowsiness and fatigue are also observed after alcohol consumption ~\cite{c40}. These changes in eye movement and facial expressions can be analyzed using face videos of drunk and sober people. Affective analysis has been successfully used to detect complex mental states such as depression, psychological distress, and truthfulness ~\cite{c38} ~\cite{c39}.

\begin{figure}[b]
\centering
\includegraphics[width=\linewidth]{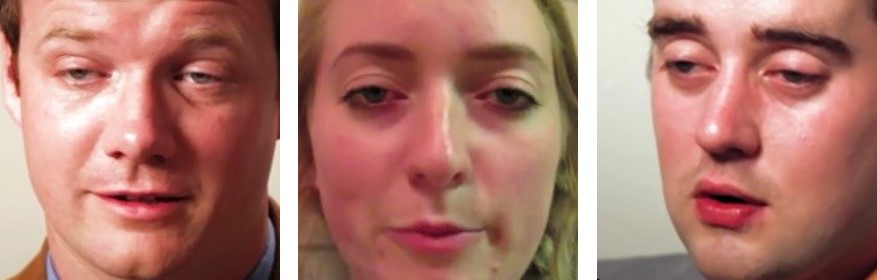}
\caption{Symptoms of fatigue observed in faces from the intoxicated category in the DIF dataset.}

\end{figure}

A review of facial expression and emotion recognition techniques is necessary to implement a system for detecting intoxication using face videos. Early attempts to parameterize facial behavior led to the development of the Facial Action Coding System (FACS) ~\cite{c7} and Facial Animation Parameters (FAPs) ~\cite{c8}. The Cohn-Kanade Database ~\cite{c6} and MMI Facial Expression Database ~\cite{c9} provide videos of facial expressions annotated with action units.

Recent submissions to the Emotion Recognition In The Wild Challenge by Dhall et al. ~\cite{c14} have focused on deep learning-based techniques for affective analysis. In EmotiW 2015, Kahou et al. ~\cite{c20} used Recurrent Neural Network (RNN) combined with a CNN for modeling temporal features. In EmotiW 2016, Fan et al. \cite{fan2016video} used RNN in combination with 3D convolutional networks to encode appearance and motion information, achieving state-of-the-art performance. Using a CNN to extract features for a sequence and classifying it with RNN performs well on emotion recognition tasks. A similar approach can be used for drunk person identification. The OpenFace toolkit ~\cite{c28} by Baltrusaitis et al. can be used to extract features related to eye gaze ~\cite{c30}, facial landmarks ~\cite{c29} and facial poses. 

In 2011, Schuller et al.\ \cite{schuller2011interspeech} organized the speaker state challenge, in which a subtask was to predict if a speaker is intoxicated. The database used in the challenge is that of Schiel et al. \cite{schiel2008alc}. Joshi et al.\ \cite{joshi2016computational} proposed a pipeline for predicting drunk texting using n-grams in Tweets. They found that drunk Tweets can be classified with an accuracy of 78.1\%. Maity et al.\ \cite{maity2018understanding} analyzed the Twitter profiles of drunk and sober users on the basis of the Tweets.

\section{Dataset Collection}
Our work consists of the collection, processing, and analysis of a new dataset of intoxicated and sober videos. We present the audio-visual database - DIF for drunk/intoxicated person identification. Figure \ref{DIFFrames} shows the cropped face frames from the intoxicated and sober categories. The dataset\footnote{\url{https://sites.google.com/view/difproject/home}}, meta-data and features will be made publicly available. 

It is non-trivial to record videos of an intoxicated driver whereas collecting data from the social networks is much easier and it could be used for intoxicated driver identification. It is interesting to note that there are channels on websites such as YouTube, where users record themselves in intoxicated states. These videos include interviews, reviews, reaction videos, video blogs, and short films. We use search queries such as `drunk reactions', `drunk review', `drunk challenge' etc. on YouTube.com and Periscope (\url{www.pscp.tv}) to obtain the videos of drunk people. Similarly, for the sober category, we collect several reaction videos from YouTube

As the downloaded videos were recorded in unconstrained real-world conditions, our database represents intoxicated and sober people `in the wild'. We use the video title and caption given on the website to assign class labels. In some cases, the subject labeled as `drunk' might only be slightly intoxicated and not exhibit significant changes in behavior. In these cases, the drunk class labels are considered as weak labels. Similarly, sober class labels are also weak labels. It is important to note that as with any other database, that has been collected from popular social networking platforms, there is always a small possibility that some videos may have been uploaded as they are extremely humorous and hence, may not represent subtle intoxication states. 

In total 78 videos are in the sober category with an average length of 10 minutes. The drunk category contains 91 videos with an average length of 6 minutes. The sober category consists of  78 unique subjects (35 males and 43 females). The drunk category consists of 88 unique subjects (55 males and 33 females).

\begin{figure}[b]
  \centering
  \includegraphics[width=80mm]{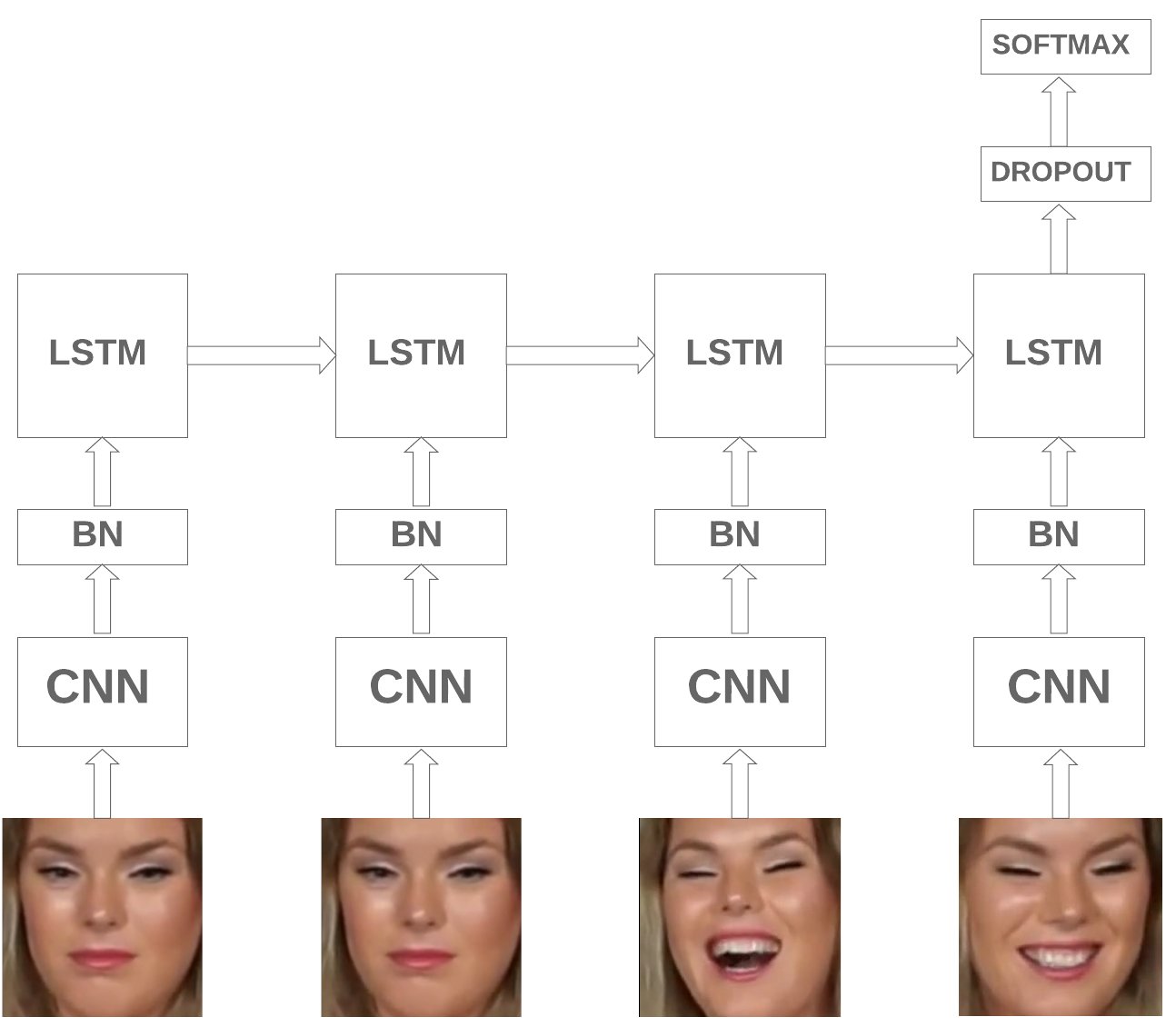}
   \caption{The CNN-RNN architecture for visual baseline. Here BN stands for Batch Normalization}

   \label{CNN-RNN}
\end{figure}

We process these videos using the pyannote-video library ~\cite{c31}. As the videos are long in duration and there are different camera shots, we perform shot detection on the video and process each shot separately in subsequent stages. Next, we perform face detection and tracking ~\cite{c28} on each shot to extract bounding boxes for faces present in each frame of the video. Based on the face tracks, we perform the clustering to extract the unique subject ids. Clustering is performed as some videos may have multiple subjects which can not be differentiated while detection and tracking. Using these ids and bounding boxes, we crop the tracked faces from the video for creating the final data sample. Hence, we obtain a set of face videos where each video contains tracked faces of an intoxicating or sober person. We also perform face alignment on each frame of the video using the OpenFace toolkit ~\cite{c28}. Sample duration was fixed at 10 seconds. This gave a total of 4,658 and 1,769 face videos for the intoxicated and sober categories, respectively.

\begin{figure*}[!htbp]
  \centering
  \includegraphics[width=178mm]{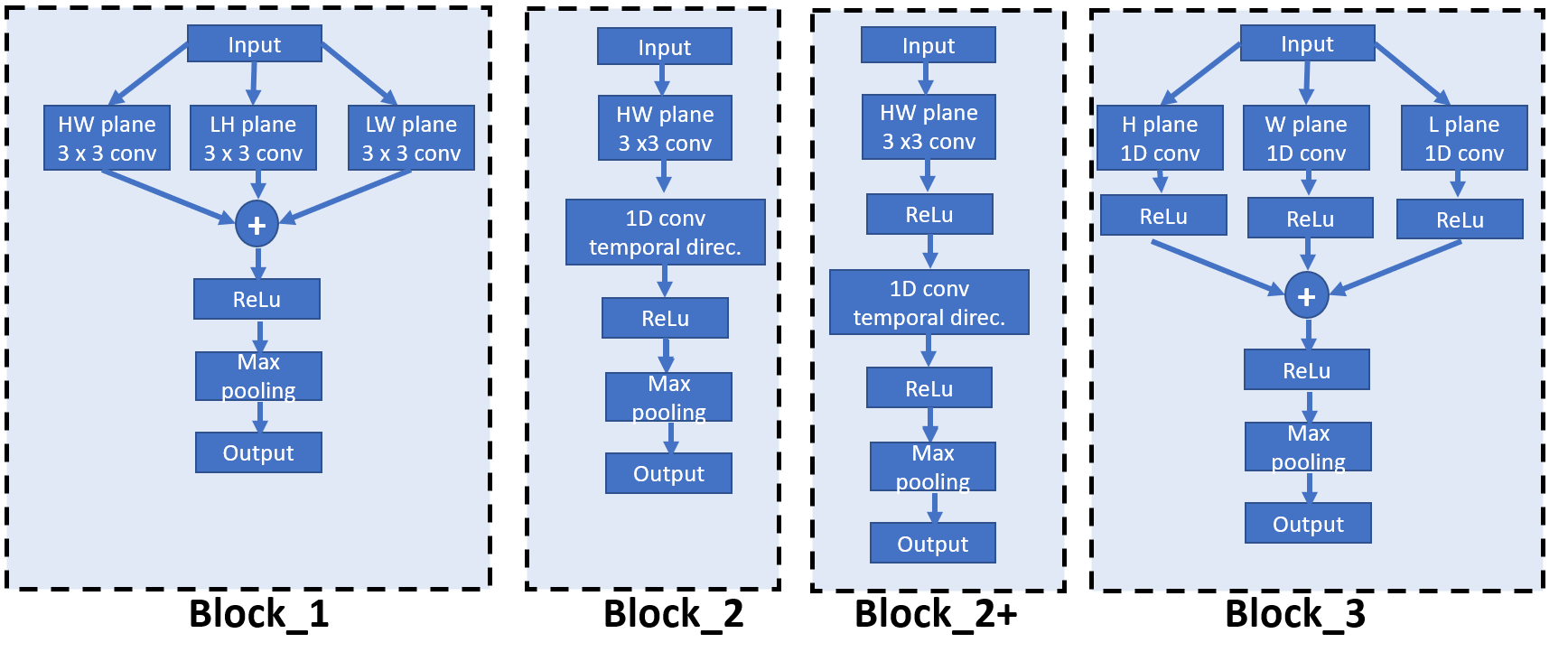}
   \caption{The figure above shows the different architecture of the 3D convolution blocks. Note that Block\_2+ has two ReLU as compared to Block\_2. Please refer to Section \ref{3DCNNText} for details.}

   \label{3DCNNS}
\end{figure*}

\section{Method}
The proposed automatic intoxicated state detector is divided into three sub-systems: the facial model, the audio model, and the ensemble strategy. The three parts are described in detail below.

\subsection{Visual Baseline}
For the visual analysis, facial appearance based on facial texture and its variation over time is important information for intoxication detection. Hence, two different neural network architectures: CNN-Recurrent Neural Network (CNN-RNN) \cite{fan2016video} and 3D CNN \cite{tran2015learning}, are utilized to extract spatio-temporal features from a video.

\subsubsection{CNN-RNN}
Fan et al.\ \cite{fan2016video} proposed the CNN-RNN based architecture for emotion recognition. The CNN based encoding extracts a discriminative representation for the aligned faces and RNN encodes the temporal change across the frames in a sample. Inspired by them, our first visual baseline shown in Figure \ref{CNN-RNN} is computed by first extracting features from pre-trained CNN and then passing the sequence of features to single layer LSTM. We used two networks pre-trained on face data: VGG-face \cite{cvgg} and AlexNet based network trained on RAF-DB \cite{li2017reliable} database. VGG-face based face representation has been successfully used in various face tasks \cite{fan2016video}. The choice of using the features from the AlexNet based model trained on the RAF-DB is that the RAF-DB is a facial expressions based database and hence, the feature extracted from the pre-trained network will be more aware of the facial deformations generated due to facial expressions.

As the difference across two neighboring frames is small in a face video and for computational efficiency, alternate frames are selected and fed into the pre-trained network to extract the features. From the  VGG-face network, fc7 layer features are extracted. During the experiments, it is observed that VGG-face based representation gave better results as compared to the RAF-DB based.

\subsubsection{3D Convolution and Variants} \label{3DCNNText}

The second visual analysis baseline computation is based on the 3D CNN. For extracting the spatio-temporal information, Tran et al.\ \cite{tran2015learning} extended the regular 2D convolutions with 3D convolutions. This is based on 3 dimensional kernels. For the basic network configuration, our 3D CNN based network contains four 3D convolution blocks. It is observed that the 3D convolution kernels based networks have large number of parameters and are computationally expensive. Further, we experiment with simple variations of the 3D convolution block. The motivation is to reduce the number of parameters, while keeping the prediction performance similar to the regular 3D convolution block. The proposed change involves replacing 3D convolution with forms of 2D convolutions and 1D convolutions. Later in the experimental section, we show that the number of parameters can be reduced without compromising the prediction performance. Figure \ref{3DCNNS} shows the 3D convolution block variants, which are discussed in detail below:

\subsubsection{Block\_1 - Three 2D Convolutions}

In this 3D convolution block variant, we divide the input volume into three orthogonal planes. Consider the input stack of frames (video) as a volume with dimensions $W \times H \times L$, where $W$ and $H$ are the spatial resolution i.e. width and height of the input video and $L$ is the number of frames. This block takes each of the three input planes, $HW, LH, LW$, separately as input and computes 2D convolutions. Then the plane-wise outputs are added together, and the ReLU activation function is applied. Finally, max pooling is applied to compute the block output. 

\subsubsection{Block\_2 - 2D + 1D Convolutions}

In this variant of the 3D convolution block, instead of treating the planes individually, as in the case of Block\_1, 2D convolutions are applied to the $HW$ of the input channels. For temporal pooling, a 1D convolution is applied to the output of the 2D convolutions from the earlier step. Then, ReLU activation function is applied followed by max pooling. In Block\_2, there are two further design choices: whether to apply the non-linearity applied twice, between the 2D and 1D convolutions and afterwards or just once, i.e. after the 1D convolutions. The rationale behind adding ReLU based non-linearity between the 2D and 1D convolutions is to increase the number of non-linearities, which may improve the performance of the network. The same is observed in the the performance of the network in the experimental section. The block with ReLU between 2D and 1D convolutions is referred to as Block\_2+.

\subsubsection{Block\_3 - Three 1D Convolutions}
In this variation, we propose to apply 1D convolutions individually in the $H, W \& L$ directions. This is followed by computing a ReLU for each dimension's output. Further, the dimension-wise outputs are added together and a ReLU activation function is applied again. Lastly, max pooling is applied to obtain the output. The idea here is that adding more non-linearities may improve the performance, though this is not observed in the experimental section.

\begin{figure}[t]
  \centering
    \includegraphics[width=90mm,height=117mm]{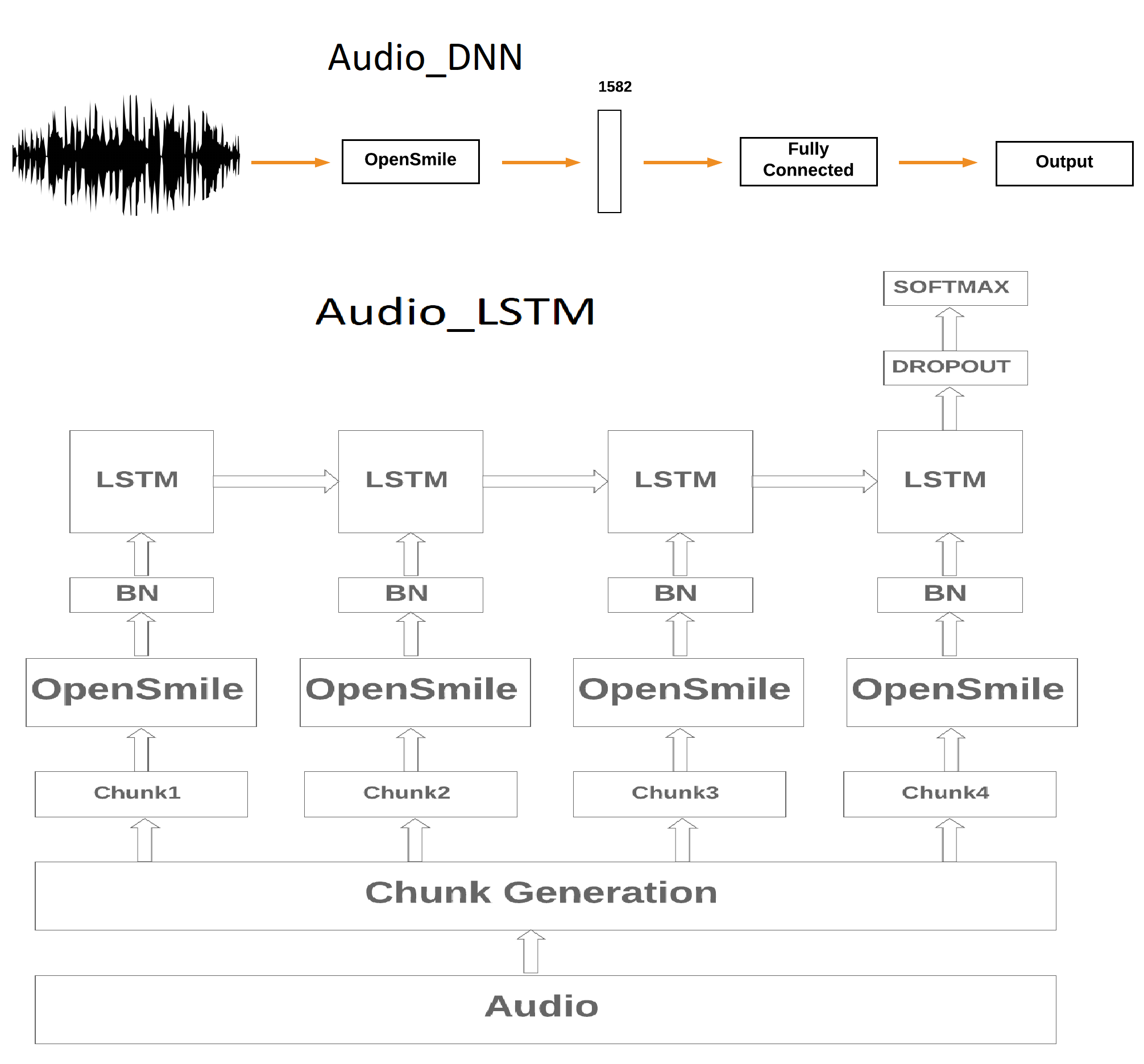}
   \caption{The two audio based networks (Audio\_DNN and Audio\_LSTM) are shown in the figure here. For details please refer to Section \ref{Audio}.}
   
   \label{AudioArchitectures}
\end{figure}

\subsection{Audio Baseline} \label{Audio}
Now let us discuss the audio modality based method. For a given sample's audio, the features are computed with the OpenSmile library \cite{eyben2010opensmile}. A set of four features are computed, these are fundamental frequency, loudness, intensity, and mel-frequency cepstral coefficients. As the first audio baseline, we train a two-layer Perceptron with ReLU activation on the OpenSmile generated feature, using batch normalization and dropout. Two different architectures are fine-tuned on the validation set. In the first architecture, the dimensions of the first and second layer are 512 and 256 respectively, while the second one consists of 256 and 128 dimensions respectively. For ease, we call this network Audio\_DNN.

We also trained a Long Short Term Memory (LSTM) network using the audio features. Based on if the window (segment) from which the OpenSmile based features are extracted is of fixed size or is variable, we tested with two variants. For the fixed size, we empirically chose audio segments of 75 ms each with an overlap of 30ms. As the duration of each sample in our database varies, the number of audio segments that will be given input to LSTM is fixed (87 segments) based on the shortest sample in the database. For the variable segment size, the number of audio segments that will be given input to LSTM is the same as that of the fixed segment size, however, the length of each segment is based on the total duration of a sample. We call this approach as Audio\_LSTM. Both the approaches Audio\_DNN and Audio\_LSTM are shown in Figure \ref{AudioArchitectures}.

\subsection{Ensemble Strategy}
In the previous sections, we describe the two models, which are the facial model and the audio model. To more adequately integrate the two modalities, we used a decision ensemble approach to promote the performance. Specifically, each model will generate a probability of the class-Drunk. The final probability $\overline{p}$ is formulated as follows,
\begin{equation}
  \overline{p}=\sum_{i=0}^{m}w_i.p_i
\end{equation}
where $w_i$ and $p_i$ are the weight and the probability by the \textit{i}th model. 
Experiments are done on average and weighted ensemble strategies. In the average ensemble, all weights are taken equal and in the second one, weights corresponds to the accuracy of the individual model on validation set.  

\section{Experiments}

We perform the face embedding and clustering to ensure that the identities present in the train, validation, and test set are different. Face recognition model based on ResNet is used to extract face embedding (\url{github.com/davisking/dlib-models}). Hierarchical agglomerative clustering is used to assign face identities. However, the face clustering procedure is not 100\% accurate and there may be some identity overlap between the sets. Subject independent data partitioning helps in accessing the generalizability performance of the methods. During the training, we save the model with the highest validation accuracy and use it for testing. The DIF database is divided into a Train set consisting of 4540 Audio-Visual (AV) samples (1045 sober and 3495 intoxicated), a Validation set consisting of 642 AV samples (321 sober and 321 intoxicated) and a Test set consisting of 948 AV samples (306 sober and 642 intoxicated). Note that the human performance on the Test set is 69.7\%.

\subsection{CNN-RNN}
In Table \ref{CNN_RNN}, we compare the performance of the CNN-RNN architecture for different input features. It is observed that the emotion features (from RAF-DB) do not perform as well as the VGG-face features. We argue that the VGG-face based features are more generalized for face analysis as the pre-trained network has learnt from over a million sample images. In comparison the RAF-DB has 26K images only. 

Further, it is also plausible that the large difference in performance may also be due to the difference in ethnicity. RAF-DB mainly contains Asian subjects, compared to this there is no such constraint in the VGG-face. This raises interesting discussion for future for transfer learning across data from different ethnicity. For further comparison, we use the VGG-face based CNN-RNN network.

 \begin{table}[t]
 \caption{Comparison of face representation extracted from pre-trained network for CNN-RNN on the Validation set.} 
 \label{CNN_RNN}
 \centering
    \begin{tabular}[t]{ |p{3cm}|p{3cm}|  }
        \hline
        \multicolumn{2}{|c|}{\textbf{CNN-RNN} } \\
        \hline
        Pre-trained Model & Validation Accuracy\\
        \hline
        VGG-Face   & 78.12\%\\
        \hline
        Emotion RAF & 62.02\%\\
        \hline
    \end{tabular}
\end{table}

\subsection{3D Convolution and Variants}

The 3D convolution block and its variants are shown in Figure \ref{3DCNNS}. The architectures are trained using the Adam optimizer, cross entropy loss and a batch size of 2. Classification accuracy on the validation set based comparison of the four blocks is mentioned in Table \ref{tab:3DCNN}. Along with the accuracy, the number of parameters and the factor of decrease in the number of parameters with respect to fully 3D CNN i.e. containing a 3D convolution block only is also mentioned. In all experiments, the number of 3D convolution (and variants) blocks is fixed to four. Replacing the later two 3D convolution blocks with two Block\_1 increases the accuracy slightly. However, please note that the number of parameters still remains the same.

When two Block\_2 are replaced in the fully 3D convolution network, we observe a drop in the accuracy by \textasciitilde3.8\% and the number of parameters are reduced by 1.8 times compared to the fully 3D convolution. This motivated us to add a ReLU between the 2D and 1D convolutions (Block\_2+). Two Block\_2+ at the end of the network resulted in an increase in performance of \textasciitilde3.57\% as compared to fully 3D convolution with the decrease in number of parameters (1.8 times). Later we tried by replacing three 3D convolution blocks with three Block\_2. Here, contrary to the decrease in accuracy with only two Block\_2 at the end, the performance increased by \textasciitilde2.7\%. It is interesting to note that three Block\_2's performance is still less than two Block\_2+. This stems the importance of adding non-linearity between the 2D and 1D blocks. With three Block\_2 at the end i.e. the network has 1 3D convolution block followed by three Block\_2, the number of parameters decreased by 2.1 times.  
Similarly, when three Block\_2+ replace the later three 3D convolution blocks the performance jumps to 78.06\%. We observed that with Block\_3, the performance was inferior as compared to all other configurations, however, the number of parameters are least in comparison. The drop in performance can be attributed to the loss in spatial information as $H$ and $W$ are treated separately with 1D convolutions. 

For fusion, we consider the best performance configuration, i.e. network with one 3D convolution and three Block\_2+. Note that in the all blocks above, the kernel size = 3, stride = 2 and padding = 1. The baseline code and data will be shared publicly.

\begin{table}[t]
\centering
\caption{Comparison of the 3D convolution block variants. Note that the last column shows the decrease in the amount of parameters are compared to a traditional four 3D convolution block architecture (Row 3). For Two Block\_1 the number of parameters is same as the default 3D convolution architecture.} \label{tab:3DCNN}
    \begin{tabular}{ |p{2.5cm}|p{1.5cm}|p{1.5cm}|p{1.5cm}|  }
        \hline
        \multicolumn{4}{|c|}{\textbf{3D CONVOLUTION and VARIANTS} } \\
        \hline
        \textbf{Model Name} & Test     & Conv.       & Parameter     \\
                     & Accuracy & Parameters  & decrease factor \\
        \hline
        Fully 3D CNN   & 73.84\% & 7.8e5
        & -----------\\
        \hline
        Two Block\_1    & 74.05\%
        & 7.8e5 & -----------\\
        \hline
        Two Block\_2    & 70.04\%
        & 4.37e5 & 1.8 times\\
        \hline
        Two Block\_2+   & 77.43\% & 4.37e5 & 1.8 times\\
        \hline
        Three Block\_2   & 76.58\%
        & 3.75e5 & 2.1 times\\
        \hline
        Three Block\_2+   & \textbf{78.06\%}
        & \textbf{3.75e5} & \textbf{2.1 times}\\
        \hline
        Two Block\_3
        & 69.20\%  & 3.4e5 & 2.3 times
        \\
        \hline
    \end{tabular}

\end{table}

\begin{table}[b]
\caption{Comparison of the Audio\_DNN configurations.} 
\label{AudioDNN}
    \begin{tabular}{ |p{3.5cm}|p{3.5cm}|  }
        \hline
        \multicolumn{2}{|c|}{\textbf{Audio} } \\
        \hline
        \textbf{Layer Dimension} &  \textbf{Validation Accuracy}\\
        \hline
        256-128   & 85.23\%\\
        \hline
         512-256 & 88.51\%\\
        \hline
    \end{tabular}
\end{table}

The run time complexity of Block\_2+ can be improved by adopting the orthogonal plane strategy of the popular Local Binary Pattern Three Orthogonal Planes feature descriptor \cite{zhao2007dynamic}. Instead of convolving over the whole volume, orthogonal planes from the input volume can be used. 

\subsection{Audio}
In the case of Audio\_DNN, we tried two configurations as shown in Table \ref{AudioDNN}. The 512-256 performs better by a margin of \textasciitilde3.2\%. In the case of Audio\_LSTM, the classification accuracy of the fixed segment (chunk) size network is 74.40\%. In the case of the variable segment size, Audio\_LSTM the classification accuracy improves to 82.80\% on the validation set.

\subsection{Ensemble}

A comparison between the ensemble techniques is shown in Table \ref{Ensemble}. The metrics used here are classification accuracy, precision, and recall. These results are generated on the test set.
It is interesting to note that the performance of the audio model is better than the other models. Note that here for the 3D CNN model, we consider a 3D convolution block followed by three Block\_2+. On the test set, the performance of RNN-LSTM and 3D CNN is similar. A Weighted ensemble increases the performance marginally by \textasciitilde1\%. It is clear that more sophisticated fusion techniques need to be investigated in the future for this task. 
We qualitatively investigated the results and found that there are a few samples, where the vision-based networks give an incorrect prediction for the intoxicated subject, who have less facial movement. Though the classification prediction is correct with the audio network. In other cases, there are some samples, where the subject speaks quite less. Here, the audio network misclassifies, while both vision networks correctly classify the sample as intoxicated. The weighted fusion architecture correctly classified a few samples from both the cases mentioned above.   

\begin{table}[t]
\caption{Comparison in terms of accuracy, precision and recall of the individual modalities and the ensemble.} 
\label{Ensemble}
    \begin{tabular}{ |p{3.0cm}|p{1.5cm}|p{1.5cm}|p{1.0cm}|  }
        \hline
        \multicolumn{4}{|c|}{\textbf{ ENSEMBLE TEST SET EVALUATION} } \\
        \hline
        \textbf{Model Name} & \textbf{Accuracy} & \textbf{Precision} & \textbf{Recall}\\
        \hline
        VGG-LSTM   & 76.37\% & 0.78 & 0.98\\
        \hline
        3D CNN (Block\_2+)  & 77.42\%\textbf{} & 0.79 & 0.90\\
        \hline
        Audio  & 87.55\%\textbf{} & 0.85 & 0.98\\
        \hline
        Ensemble (Average)  & 87.55\%\textbf{} & 0.85 & 0.98\\
        \hline
        \textbf{Ensemble (Weighted)}  & \textbf{88.39\%} & \textbf{0.85} & \textbf{0.99}\\
        \hline
    \end{tabular}

\end{table}

\section{Conclusion, Limitations and Future Work}

We collected the DIF database in a semi-supervised manner (weak labels from video titles) from the web, without major human intervention. The DIF database, to the best of our knowledge, is the first audio-visual database for predicting if a person is intoxicated. The baselines are based on CNN-RNN, 3D CNN, audio-based LSTM, and DNN. It is interesting to note that the audio features extracted using OpenSmile performed better than the CNN-RNN and 3D CNN approaches. Further, analysis is performed on decreasing the complexity of the 3D CNN block by converting the 3D CNN into a set of 2D and 1D convolutions. The proposed variant of the 3D Convolution, i.e., Block\_2+ improved the accuracy as well as decreased the number of parameters compared to the fully 3D baseline. The present ensemble model of CNN-RNN, 3D CNN, audio features performed well.

Rather than providing a clinically accurate database, we have shown that intoxication detection using online videos is viable even from videos collected in the wild which could be used in future to detect intoxicated driving using transfer learning. Further, research may include videos from a controlled environment with exact intoxication data for better accuracy. 

We are dealing with intoxication detection as a binary classification problem. It will be interesting to investigate intoxication detection as a continuous regression problem once more data representing different intoxication state intensities is collected. We note that since the database uses a relatively small number of source videos collected in the wild to extract face videos, there may be visual biases present in the videos of sober and drunk classes. Specifically, some drunk class videos have low resolution, poor lighting, and lower brightness compared to sober class videos. These subtle differences can help the model achieve higher accuracy without learning to discriminate between sober and drunk faces. We also note that due to resource constraints, we use a single train validation-test split of the dataset. In future work, we aim to perform 5-fold cross-validation to address any bias due to a selection of the test set.

An interesting comparison of the 3D convolution blocks would be that with 3D separable kernels. It has been noted that for, behavior analysis, head pose and eye gaze provide discriminative cues. In the future, eye gaze and head pose information will be added to the network. There is scope in exploring different fusion strategies of the audio and visual modalities. Further, to create a mobile phone-based app, it would be important to explore networks with a smaller footprint. Currently, the duration of a sample is fixed to 10 seconds, but there is scope for exploring different sample durations with both fixed and variable configurations. 

Further, there is scope for distillation based learning, video network being computationally more expensive as compared to audio-based networks, there is scope for transferring knowledge to an audio-based network from the vision-based network. As it is observed in the performance comparison of Block\_2+ of Block\_2 that extra non-linearity is helping in achieving better performance, it may be worthwhile to explore different non-linearity functions, such as swish and leaky-ReLU, etc.

\section*{Acknowledgement}
We acknowledge the support of NVIDIA for providing us TITAN Xp GPU for research purposes.

\bibliographystyle{ACM-Reference-Format}
\bibliography{sample-bibliography}

\end{document}